\documentclass[runningheads]{llncs}
\usepackage{graphicx}

\usepackage{tikz}
\usepackage{comment}
\usepackage{amsmath,amssymb} 
\usepackage{color}
\usepackage{epsfig}
\usepackage{graphicx}
\usepackage{amsmath}
\usepackage{amssymb}
\usepackage{makecell}
\usepackage{multirow, booktabs}
\usepackage{tabularx}
\usepackage{cuted}
\usepackage{capt-of}
\usepackage[accsupp]{axessibility}  
\newcommand{\eg}{\textit{e}.\textit{g}. }
\newcommand{\ie}{\textit{i}.\textit{e}. }
\newcommand{\tablestyle}[2]{\setlength{\tabcolsep}{#1}\renewcommand{\arraystretch}{#2}\centering\footnotesize}

\usepackage[pagebackref=true,breaklinks=true,letterpaper=true,colorlinks,bookmarks=false]{hyperref}

\begin{document}
\pagestyle{headings}
\mainmatter
\def\ECCVSubNumber{100}  

\title{Egocentric Activity Recognition and Localization on a 3D Map} 

\titlerunning{Egocentric Activity Recognition and 3D Localization }
\author{
  Miao Liu$^{1}$\thanks{This work was primarily done during an internship at Meta Reality Labs.} ,\;Lingni Ma$^{5}$,\; Kiran Somasundaram$^{5}$,\; Yin Li$^{2}$,\; Kristen Grauman$^{3,4}$,\; James M. Rehg$^{1}$,\; Chao Li$^{5}$}  
  
\institute{ 
    Georgia Institute of Technology\\ \and
    University of Wisconsin-Madison\\ \and
    The University of Texas at Austin \\ \and
    Meta AI\\ \and
    Meta Reality Labs\\ 
}
\authorrunning{M. Liu et al.}
\maketitle

\begin{abstract}
Given a video captured from a first person perspective and the environment context of where the video is recorded, can we recognize what the person is doing and identify where the action occurs in the 3D space? We address this challenging problem of jointly recognizing and localizing actions of a mobile user on a known 3D map from egocentric videos. To this end, we propose a novel deep probabilistic model. Our model takes the inputs of a Hierarchical Volumetric Representation (HVR) of the 3D environment and an egocentric video, infers the 3D action location as a latent variable, and recognizes the action based on the video and contextual cues surrounding its potential locations. To evaluate our model, we conduct extensive experiments on the subset of Ego4D dataset, in which both human naturalistic actions and photo-realistic 3D environment reconstructions are captured. Our method demonstrates strong results on both action recognition and 3D action localization across seen and unseen environments. We believe our work points to an exciting research direction in the intersection of egocentric vision, and 3D scene understanding.

\end{abstract}

\section{Introduction}

\begin{figure}[t]
\centering
\includegraphics[width=0.89\linewidth]{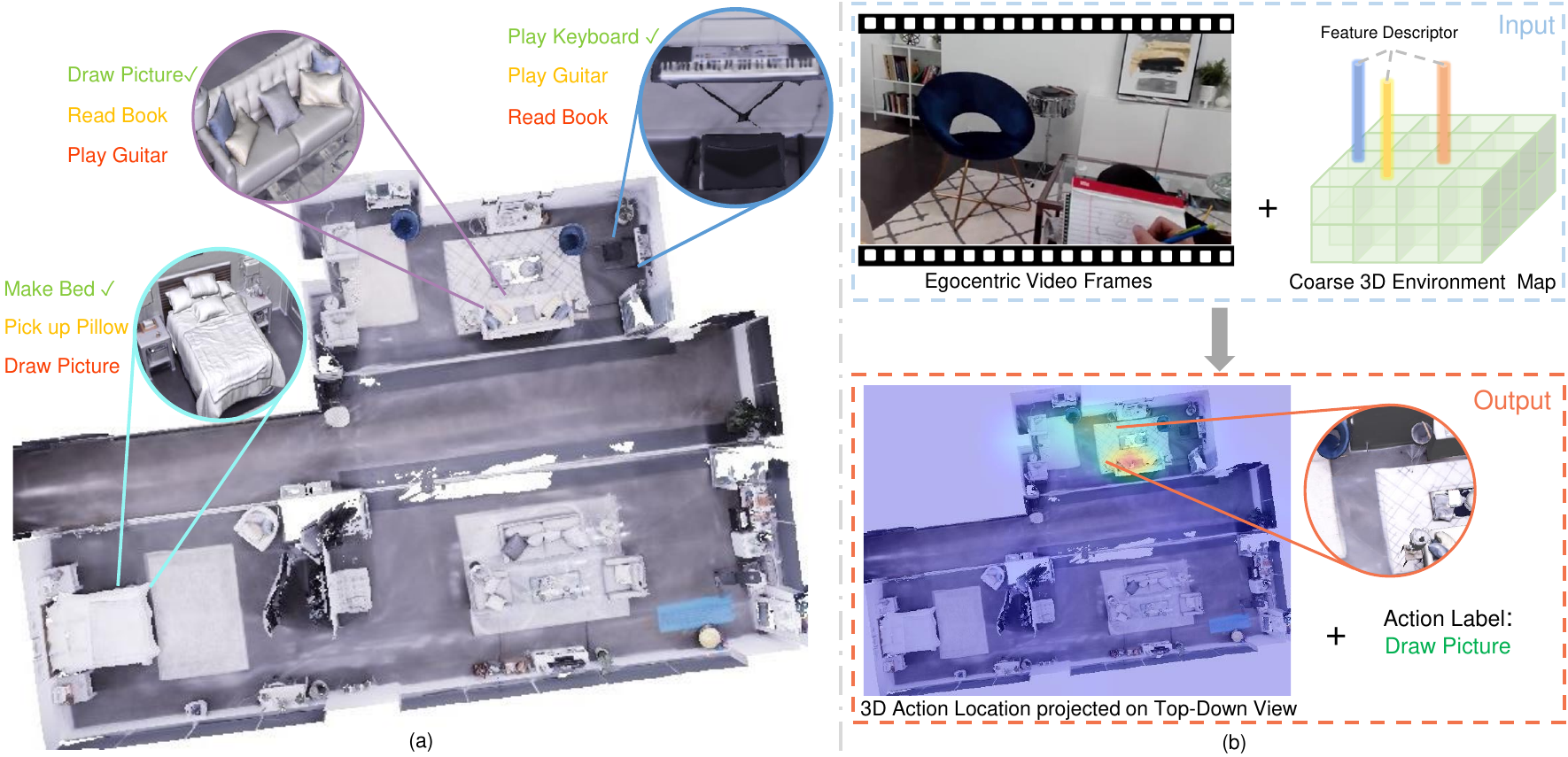}
\caption{(a) The activities of daily life take place in a 3D environment, and the semantic and spatial properties of the environment are powerful cues for activity recognition. (b) \emph{Our Proposed Task}: Given an input egocentric video sequence and a 3D volumetric representation of the environment (carrying both semantic and geometric information), our goal is to detect and localize activities, by jointly predicting the action label and location on the 3D map where it occurred.} 
\label{fig:teaser}
\end{figure}

Egocentric vision has emerged as a promising paradigm for understanding human activities in a mobile setting. Its defining characteristic is the continuous capture of first-person visual experience.
In particular, egocentric videos implicitly and naturally connect the camera wearer's activities to the relevant 3D spatial context, such as the surrounding objects and their 3D layout. While this observation has been true since the beginning of egocentric vision, it is only recently that 3D scene models that can capture this context have become readily available, due to advances in 3D scanners~\cite{sulaiman2020matterport} and Augmented Reality (AR) headsets~\cite{karthika2017hololens}. Fig.\ \ref{fig:teaser} (a) gives an example of a 3D scan of a subject's apartment in which the 3D layout of the furniture and appliances is known {\it a priori}. Given an egocentric video, our goal is to leverage this 3D map to reason about the camera wearer's activities and the 3D locations in which they are performed, \eg drawing a picture while sitting on the sofa. Such a capability could enable future context-sensitive applications in AR and Human-Robot Interaction (HRI).
This paper addresses the following research question: {\it How can we design vision models to exploit the prior knowledge of a known 3D environment for recognizing and localizing egocentric activities?} This question has not been tackled by existing works on egocentric action and activity recognition~\cite{kazakos2019epic,Zhou_2016_CVPR,ma2016going,li2015delving,pirsiavash2012detecting,moltisanti2017trespassing,Poleg_2014_CVPR,Li_2018_ECCV}. Prior works have used limited contextual cues for egocentric video analysis, such as a 2D ground plane~\cite{rhinehart2016learning} or a topological map~\cite{ego-topo}, and have focused on understanding the functions of an environment, such as the common locations at which activities occur. In contrast, this work introduces the new task of the \emph{joint recognition and 3D localization of egocentric activities given trimmed videos and a coarsely-annotated 3D environment map}. We provide a visual illustration of our problem setting in Fig.\ \ref{fig:teaser} (b). 
Two major challenges arise in our task. First, standard architectures for egocentric activity recognition are not designed to incorporate 3D scene context, requiring a new design of action recognition models and associated 3D scene representations. Second, the exact ground truth for the locations of actions in a 3D scene that is the size of an entire apartment is difficult to obtain, due to ambiguities in 2D to 3D registration. As a remedy, we leverage camera registration using structure-from-motion that yields ``noisy'' locations, which requires the model to address the uncertainty in action locations during training. 



To address the challenge of leveraging context in recognition, we develop a Hierarchical Volumetric Representation (HVR) to describe the semantic and geometric information of the 3D environment map (see Fig.\ \ref{fig:overview} (a) and Sec 3.1 for explanation). We further present a novel deep model that takes egocentric videos and our proposed 3D environment HVR as inputs, and outputs the 3D action locations and the activity classes. Our model consists of two branches. The \emph{environment branch} makes use of a 3D convolutional network to extract global environmental features from HVR. Similarly, the \emph{video branch} uses a 3D convolutional network to extract visual features from the input video. The environmental and visual features are further combined to estimate the 3D activity location, supervised by the results of camera registration. Moreover, we tackle the second challenge of noisy localization by using stochastic units to account for uncertainty. The predicted 3D activity location, in the form of a probabilistic distribution, is then used as a 3D attention map to select local environmental features relevant to the action. Finally, these local features are further fused with video features for recognition. 

Our method is trained and evaluated on the recent, freely-available Ego4D dataset~\cite{grauman2021ego4d}, which contains naturalistic egocentric videos and photo-realistic 3D scene reconstructions along with 3D static object annotations. We demonstrate strong results on action recognition and 3D action localization. Specifically, our model outperforms a strong baseline of 2D video-based action recognition methods by 4.2\% in mean class accuracy, and beats baselines on 3D action localization by 9.3\% in F1 score. Furthermore, we demonstrate that our method can generalize to unseen environments not present in the training set yet with known 3D maps and object labels. 
We believe this work provides a useful foundation for egocentric video understanding in a 3D scene context

\section{Related Work}
We first discuss the most relevant works on egocentric vision, and then review several previous efforts on human-scene interaction and 3D scene representation.

\noindent\textbf{Egocentric Vision}.\ There is a rich set of literature aiming at egocentric activity understanding. Prior works have made great progress in recognizing and anticipating egocentric actions based on 2D videos~\cite{furnari2019rulstm,shen2018egocentric,Ke_2019_CVPR,Zhou_2016_CVPR,ma2016going,li2015delving,pirsiavash2012detecting,moltisanti2017trespassing,Poleg_2014_CVPR}, and predicting gaze and locomotion~\cite{Li_2018_ECCV,li2013learning,huang2018predicting,soo2016egocentric,ng2020you2me,Zhang_2017_CVPR,park20123d,poleg2014head,serra2013hand}. Far fewer works have considered environmental factors and spatial grounding of egocentric activity. Guan et al.\ \cite{guan2020generative} and Rhinehart et al.\ \cite{Rhinehart_2017_ICCV} jointly considered trajectory forecasting and egocentric activity anticipation with online inverse reinforcement learning. The most relevant works to ours are recent efforts on learning affordances for egocentric action understanding~\cite{ego-topo,rhinehart2016learning}. Nagarajan et al.\ \cite{ego-topo} introduced a topological map environment representation for long-term activity forecasting and affordance prediction. Rhinehart et al.\ \cite{rhinehart2016learning} considered a novel problem of learning ``Action Maps" from egocentric videos. However, methods that use ground plane representations of the environment~\cite{rhinehart2016learning} or environmental functionality as the context~\cite{ego-topo} may lack the specificity provided by 3D proximity. In contrast to these prior efforts, our focus is on exploiting the geometric and semantic information in the HVR map to address our novel task of joint egocentric action recognition and 3D localization.



\noindent\textbf{Human-Scene Interaction}.\ Human-scene constraints have been proven to be effective in estimating human body model~\cite{PSI:2019,PROX:2019,PLACE:3DV:2020}. The most relevant prior works focus on understanding environment affordance. Grabner et al.\ \cite{grabner2011makes} predict object functionality by hallucinating an actor interacting with the scene. A similar idea was also explored in~\cite{jiang2013hallucinated,jiang2012learning}.Koppula et al.\ \cite{koppula2013learning} leveraged RGB-D videos to jointly recognize human activities and estimate objects affordances. Savva et al.\ \cite{savva2014scenegrok} predicted action heat maps that highlight the likelihood of an action in the scene by partitioning 3D scenes into disjoint sets of segments and learning a segment dictionary. Gupta et al.\ \cite{gupta20113d} presented a human-centric scene representation for predicting the afforded human body poses. Delaitre et al.\ \cite{delaitre2012scene,fouhey2014people} introduced a statistical descriptor of person-object interactions for object recognition and human body pose prediction. Fang et al.\ \cite{fang2018demo2vec} proposed to learn object affordances from demonstrative videos. Nagarajan et al.\ \cite{nagarajan2018grounded} proposed to use backward attention to approximate the interaction hotspots of future action. Those previous efforts were limited to the analysis of environment functionality
~\cite{jiang2013hallucinated,jiang2012learning,delaitre2012scene,fouhey2014people}, constrained human action and body pose~\cite{savva2014scenegrok}, or hand-object interaction on 2D image plane~\cite{fang2018demo2vec,nagarajan2018grounded}. In contrast, we are the first to utilize the rich geometric and semantic information of the 3D scene for naturalistic human activity recognition and 3D localization.

\noindent\textbf{3D Scene Representation}.\ Many recent works explored various 3D representations for 3D vision tasks, including 3D object detection~\cite{yang2018pixor,zhou2018voxelnet,chen2017multi} and embodied visual navigation~\cite{gordon2019splitnet,Henriques_2018_CVPR,Li_2020_CVPR}. Deep models have been developed for point clouds \cite{qi2017pointnet,wu2020pointpwc,wu2019pointconv,qi2017pointnetplusplus} with great success in object recognition, semantic segmentation, and sceneflow estimation. However, using point clouds to describe a large-scale 3D scene will result in high computational and memory cost~\cite{zhou2018voxelnet}. To address this challenge, many approaches used rasterized point clouds in a 3D voxel grid, with each voxel represented by either handcrafted features~\cite{wang2015voting,engelcke2017vote3deep,song2014sliding,song2016deep} or learning-based features~\cite{zhou2018voxelnet}. Signed-Distanced value and Chamfer Distance between 3D scene and 3D human body have also been used to enforce more plausible human-scene contact~\cite{PSI:2019,PROX:2019,PLACE:3DV:2020,liu20204d}. Building on this prior works, we utilize a 3D Hierarchical Volumetric Representation (HVR) that encodes the geometric and semantic context of a 3D scene for egocentric activity understanding.


\section{Method}
We denote a trimmed input egocentric video as $x=(x^1, ..., x^t)$ with frames $x^t$ indexed by time $t$. In addition, we assume a global 3D environment prior $e$, associated each with input video, is available at both training and inference time. $e$ is environment specific, \eg the 3D map of an apartment. Our goal is to jointly predict the action category $y$ of $x$ and the action location $r$ on the 3D map. $r$ is parameterized as a 3D saliency map, where the value of $r(w,d,h)$ represents the likelihood of action clip $x$ happening in spatial location $w,d,h$. For tractability, we associate the entire activity with a specific 3D location and do not model location change over the course of an activity. This is a valid assumption for the activities we address, such as sitting down, playing keyboards, etc.
$r$ thereby defines a proper probabilistic distribution in 3D space. 

In this section, we first introduce our proposed joint model of action recognition and 3D localization, leveraging a 3D representation of the environment. We then describe key components of our model, training and inference schema, as well as our network architecture.

\begin{figure*}[t]
\centering
\includegraphics[width=0.99\linewidth]{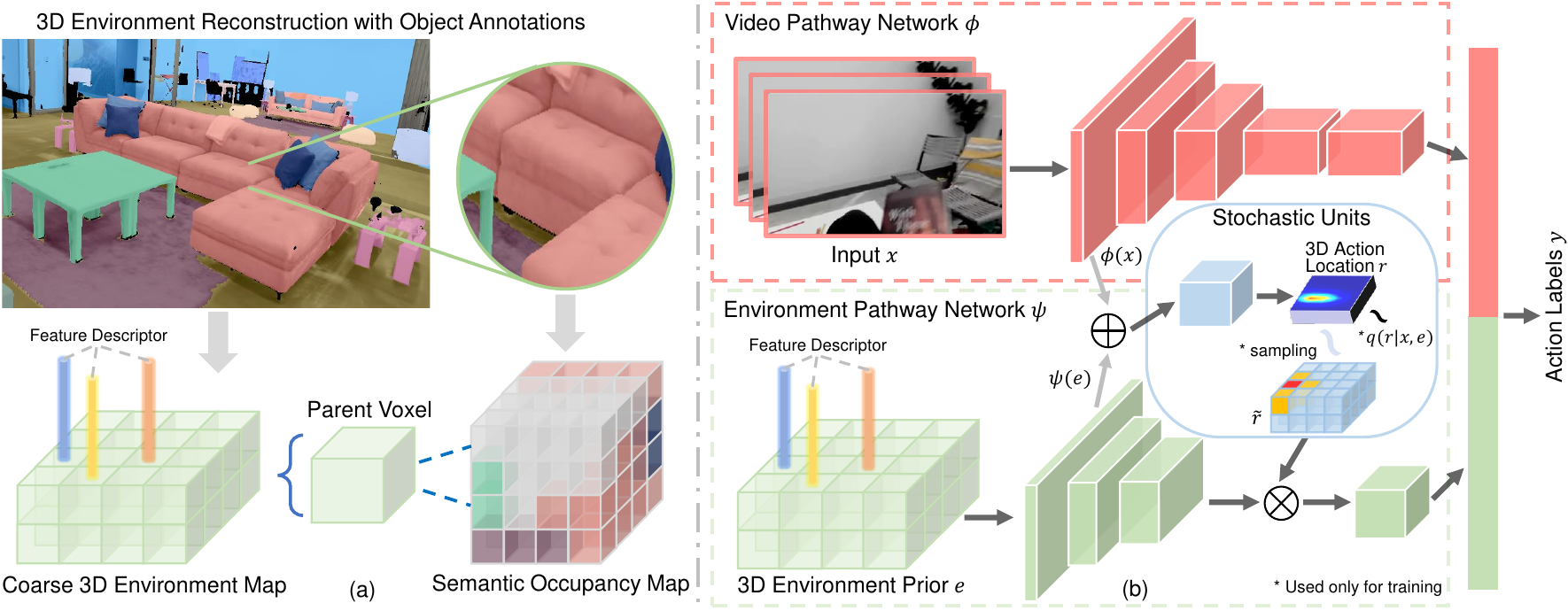}
\caption{(a) Hierarchical Volumetric Representation (HVR). We rasterize the semantic 3D environment mesh into two levels of 3D voxels. Each parent voxel corresponds to a possible action location, while the children voxels compose a semantic occupancy map that describes their parent voxel. (b) Overview of our model. Our model takes video clips $x$ and the associated 3D environment representation $e$ as inputs. We adopt an I3D backbone network $\phi$ to extract video features and a 3D convolutional network $\psi$ to extract the global environment features. We then make use of stochastic units to generate sampled action location $\tilde{r}$ for selecting local 3D environment features for action recognition. Note that $\otimes$ represents weighted average pooling, while $\oplus$ denotes concatenation along channel dimension.} 
\label{fig:overview}
\end{figure*}

\subsection{Joint Modeling with the 3D Environment Representation}

\noindent \textbf{3D Environment Representation}.\ We seek to design a representation that not only encodes the 3D geometric and semantic information of the 3D environment, but is also effective for 3D action localization and recognition. 

To this end, we introduce a Hierarchical Volumetric Representation (HVR) of the 3D environment. We provide an illustration of our method in Fig.\ \ref{fig:overview}(a). We assume the 3D environment reconstruction with object labels is given in advance as a 3D mesh  (see Sec.4 for details). We first divide the 3D mesh into $X \times Y \times Z$ parent voxels, that define all possible action locations. We then divide each parent voxel into multiple voxels at a fixed resolution $M$ and further assign an object label to each child voxel based on the object annotation. Specifically, the object label of each child voxel is determined by the majority vote of the vertices that lie inside that child voxel. Note that we only consider static objects of the entire environments and treat empty space as a specially-designated ``object" category. Therefore, the child voxels compose a semantic occupancy map that encodes both the 3D geometry and semantic meaning of the parent voxel. 
 
We further vectorize the semantic occupancy map and use the resulting vector as a feature descriptor of the parent voxel. The 3D environment representation $e$ can then be represented as a 4D tensor, with dimension $X \times Y \times Z \times (M^3)$. Note that higher resolution $M$ can better approximate the 3D shape of the environment. Our proposed HVR is thus a compact and flexible environment representation that jointly considers the 3D action location candidates, geometric and semantic information of the 3D environment.

\noindent \textbf{Joint Learning of Action Category and Action Location}.
We present an overview of our model in Fig.~\ref{fig:overview}(b). Specifically, we adopt a two-pathway network architecture. The video pathway extracts video features with an I3D backbone network $\phi(x)$, while the environment pathway extracts the global 3D environment features with a 3D convolutional network $\psi(e)$. Visual and environmental features are jointly considered for predicting the 3D action location $r$. We then adopt stochastic units to generate sampled action $\tilde{r}$ for selecting the local environment features relevant to the actions. Local environment features and video features are further fused together for activity recognition.

Our key idea is to utilize the 3D environment representation $e$ for jointly modeling the action label $y$ and 3D action location $r$ of video clip $x$. We consider the  action location $r$ as a probabilistic variable, and model the action label $y$ given input video $x$ and environment representation $e$ using a latent variable model. Therefore, the conditional probability $p(y|x,e)$ is given by: 
\begin{equation}
\label{eq:joint}
    p(y|x,e) = \int_{r} p(y|r,x,e) p(r|x,e) dr.
\end{equation}
Notably, our proposed joint model has two key components. First, $p(r|x,e)$ models the 3D action location $r$ from video input $x$ and the 3D environment representation $e$. Second, $p(y|r,x,e)$ utilizes $r$ to select a region of interest (ROI) from the environment representation $e$, and combines selected environment features with the video features from $x$ for action classification. During training, our model receives the ground truth 3D action location and action label as supervisory signals. At inference time, our model jointly predicts both the 3D action location $r$ and action label $y$. We now provide additional technical details in modeling $p(r|x,e)$ and $p(y|r,x,e)$.


\subsection{3D Action Localization}
We first introduce our 3D action localization module, defined by the conditional probability $ p(r|x,e)$. Given the video pathway features $\phi(x)$ and the environment pathway features $\psi(e)$, we learn a mapping function to predict location $r$, which is defined on a 3D grid of candidate action locations. Note that the 3D grid is defined globally over the 3D environment scan. The mapping function is composed of 3D convolution operations with parameters $w_r$ and a softmax function. Thus, $p(r|x,e)$ is given by:
\begin{equation}
\label{eq:grounding}
    p(r|x,e) = softmax(w_r^T(\phi(x) \oplus \psi(e))),
\end{equation}
where $\oplus$ denotes concatenation along the channel dimension. Therefore, the resulting action location $r$ is a proper probabilistic distribution normalized in 3D space, and $r(w,d,h)$ can be considered as the expectation of video clip $x$ happening in the spatial location $(w,d,h)$ of the 3D environment.

In practice, we do not have access to the precise ground truth 3D action location and must rely on camera registration results as a proxy. Using a categorical distribution for $p(r|x,e)$ thus models the ambiguity of 2D to 3D registration. We follow~\cite{li2020eye,liu2019forecasting} to adopt stochastic units in our model. Specifically, we follow the Gumbel-Softmax and reparameterization trick from~\cite{jang2016categorical,maddison2016concrete} to adopt the following differentiable sampling mechanism:
\begin{equation}
\label{eq:sample}
\tilde{r}_{w,d,h} \sim \frac{\exp ((\log r_{w,d,h} + G_{w,d,h})/\theta)}{\sum_{w,d,h} \exp ((\log r_{w,d,h} + G_{w,d,h})/\theta)},
\end{equation}
where $G$ is a Gumbel Distribution for sampling from a discrete distribution. This Gumbel-Softmax trick produces a ``soft'' sample that allows the gradients propagation to video pathway network $\phi$ and environment pathway network $\psi$. $\theta$ is the temperature parameter that controls the shape of the soft sample distribution. We set $\theta=2$ for our model. Notably, the expectation of sampled 3D action location $E[\tilde{r}]$ can be modeled by the distribution $p(r|x, e)$ using Eq.~\ref{eq:grounding}.

\subsection{Action Recognition with Environment Prior}
Our model further models $p(y|r,x,e)$ with a mapping function $f(\tilde{r},x,e)$ that jointly considers action location $r$, video input $x$ and 3D environment representation $e$ for action recognition. Formally, the conditional probability $p(y|r,x,e)$ can be modeled as:
{
\begin{align}
\label{eq:recog}
p(y|r,x,e) = f(\tilde{r},x,e) = softmax(w_p^T\Sigma  (\phi(x) \oplus (\tilde{r}\otimes \psi(e)) )),
\end{align}
}%
where $\oplus$ denotes concatenation along channel dimension, and $\otimes$ denotes the element-wise multiplication. Specifically, our method uses the sampled action location $\tilde{r}$ for selectively aggregating environment features $\psi(e)$ and combines the aggregated environment features with video features $\phi(x) $ for action recognition. $\Sigma$ denotes the average pooling operation that maps 3D feature to 2D feature, and $w_p$ denotes the parameters of the linear classifier that maps feature vector to action prediction logits.

\subsection{Training and Inference}
We now present our training and inference schema. At training time, we assume a prior distribution of action location $q(r|x,e)$ is given as a supervisory signal. $q(r|x,e)$ is obtained by registering the egocentric camera into the 3D environment (see more details in Sec.4). Note that we factorize $p(r|x,e)$ as latent variables, and based on the Evidence Lower Bound (ELBO), the resulting deep latent variable model has the following loss function:
{
\begin{align}
\label{eq:loss}
\mathcal{L} &=-\sum_{r} \log p(y|r,x,e) + KL[p(r|x,e) ||q(r|x,e)],
\end{align}}%
where the first term is the cross entropy loss for action classification and the second term is the KL-Divergence that matches the predicted 3D action location distribution $p(r|x,e)$ to the prior distribution $q(r|x,e)$. During training, a single 3D action location sample $\tilde{r}$ for each input within the mini-batch will be drawn.

Theoretically, our model should sample $\tilde{r}$ from the same input multiple times and take average of the predictions at inference time. To avoid such dense sampling for high dimensional video input, we choose to directly plug in the deterministic action location $r$ in Eq.~\ref{eq:recog}. Note that the recognition function $f$ is composed of a linear mapping function and a softmax function, and therefore is convex. Further, $\tilde{r}$ is sampled from the probabilistic distribution of 3D action location $r$, similar to the formulation in~\cite{Li_2018_ECCV,liu2019forecasting}, $r$ is thus the expectation of $\tilde{r}$. By Jensen’s Inequality, we have:
\begin{equation}
\label{eq:jesen}
E[f(\tilde{r},x,e)] \ge f(E[\tilde{r}],x,e) = f(r,x,e). 
\end{equation}
\noindent That being said, $f(r,x,e)$ provides an empirical lower bound of $E[f(\tilde{r},x,e)]$, and therefore provides a valid approximation of dense sampling.

\subsection{Network Architecture}
For the video pathway, we adopt the I3D-Res50 network~\cite{carreira2017quo,wang2018non} pre-trained on Kinetics as the backbone. For the environment pathway, we make use of a lightweight network (denoted as EnvNet), which has four 3D convolutional operations. The video features from the 3rd convolutional block of I3D-Res50 and the environment features after the 2nd 3D convolutional operation in EnvNet are concatenated for 3D action location prediction. We then use 3D max pooling operations to match the size of action location map to the size of the feature map of the 4th convolution of EnvNet for the weighted pooling in Eq.\ref{eq:grounding}. More implementation details can be found in our supplement.

\section{Experiments and Results}

\subsection{Dataset and Benchmarks}
\noindent \textbf{Datasets}. Note that existing egocentric video datasets (EGTEA~\cite{li2020eye}, and EPIC-Kitchens~\cite{damen2020epic} etc.) did not explicitly capture the 3D environment. We follow~\cite{rhinehart2016learning} to run ORB-SLAM on EGTEA and EPIC-Kitchens. However, less than 30\% of frames can be registered, and the quality of the reconstructed point cloud is unsatisfactory. Our empirical finding is that existing visual SfM methods can not address the naturalistic egocentric videos. In contrast, the newly-developed Ego4D~\cite{grauman2021ego4d} dataset has a subset that includes egocentric videos, high-quality 3D environment reconstructions, and 3D static objects annotation.


The subset captures 34 different indoor activities from 3 real-world living rooms, resulting in $6868$ action clips. Similar to~\cite{damen2020epic}, we consider both \emph{seen} and \emph{unseen} environment splits. In the seen environment split, each environment is seen in both training and testing sets ($5163$ instances for training, and $1705$ instances for testing). In the unseen split, all sequences from the same environment are either in training or testing ($4392$ instances for training, and $2476$ instances for testing). As discussed in~\cite{grauman2021ego4d}, the photo-realistic 3D reconstruction of the environment is obtained from the state-of-the-art dense reconstruction system~\cite{replica19arxiv}. Furthermore, the static 3D object meshes are annotated by painting an semantic label over the mesh polygons. The annotation includes 35 object categories plus a background class label. It is worthy noting that the static object annotations can be automated with the state-of-the-art 3D object detection algorithms.


\noindent \textbf{Prior Distribution of 3D Activity Location}.
To obtain the ground truth of the activity location for each trimmed activity video clip, we first register the egocentric camera in the 3D environment using a RANSAC based feature matching method. Specifically, we first build a base map from the monochrome camera streams for 3D environment reconstruction using Structure from Motion~\cite{frahm2010building,schoenberger2016sfm}. The pre-built base map is a dense point cloud associated with 3D feature points. We then estimate the camera pose of the video frame using active search~\cite{sattler2012improving}. Note that registering the 2D egocentric video frames in a 3D environment is fundamentally challenging, due to the drastic head rotation, featureless surfaces, and changing illumination. Therefore, we only consider the key frame camera registration, where enough inliers were matched with RANSAC. As introduced in Sec.3, the action location is defined as a probabilistic distribution in 3D space. Thus, we map the key frame camera location into the index of the 3D action location tensor, with its value representing the likelihood of the given action happening in the corresponding parent voxel. To account for the uncertainty of 2D to 3D camera registration, we further enforce a Guassian distribution to generate the final 3D action location ground truth.


\noindent \textbf{Evaluation Metrics}. For all experiments, we We follow~\cite{damen2020epic,li2020eye} to evaluate the performance of both action recognition using both Mean Class Accuracy and Top-1 Accuracy. As for 3D Action Localizatio, we consider 3D action localization as binary classification over the regular 3D grids. Therefore, we report the Precision, Recall, and F1 score on a downsampled 3D heatmap ($\times$4 in X, Y direction, and $\times$2 in Z direction) as in~\cite{Li_2018_ECCV}.

\begin{table}[t]
\centering
\caption{Comparison with other forms of environment context. Our Hierarchical Volumetric Representation (HVR) outperforms other methods by a significant margin on both action recognition and 3D action localization. The best results are highlighted with \textbf{boldface}, and the second-best results are \underline{underlined}.}
{
\tablestyle{8pt}{1.0}
\begin{tabular}{c|cc|ccc}
\toprule
\multicolumn{1}{c|}{\multirow{2}{*}{Method}}          
&\multicolumn{2}{c|}{Action Recognition} &\multicolumn{3}{c}{3D Action Localization} \\ \cline{2-6}
 & Mean Cls Acc &Top-1 Acc & Prec & Recall & F1  \\ \hline 
I3D-Res50    &37.48 &55.15 &8.14  &\underline{38.73}   &13.45 \\
I3D+Obj        &37.66  &55.11    &10.04  &35.08 &15.61\\
I3D+2DGround        &38.69  &55.37   &10.88  &36.19  &16.73\\
I3D+SemVoxel        &39.23 &\underline{56.07}    &11.26  &{\fontseries{b}\selectfont 38.77} &\underline{17.45} \\ 
I3D+Affordance       &\underline{39.95} &55.82 &\underline{11.55} &35.35 &17.41 \\
Ours(HVR)        & {\fontseries{b}\selectfont 41.64} & {\fontseries{b}\selectfont 56.94} & {\fontseries{b}\selectfont 16.71} &35.55 &{\fontseries{b}\selectfont 22.73} \\ 
\bottomrule
\end{tabular}}
\label{table:3DEnv}
\end{table}

\subsection{Action Understanding in Seen Environments} 
Our method is the first to utilize the 3D environment information for egocentric action recognition and 3D localization. Previous works have considered various environment contexts for other tasks, including 3D object detection, affordance prediction etc. Therefore, we adapt previous proposed contextual cues into our proposed joint model and design the following strong baselines (see supplementary material for the details of baseline models):

\noindent \textbullet\ \textbf{I3D-Res50} refers to the backbone network from~\cite{wang2018non}. We also use the network feature from I3D-Res50 for 3D action localization by adopting the KL loss.

\noindent \textbullet\ \textbf{I3D+Obj} uses object detection results from a pre-trained object detector\ \cite{wu2019detectron2} as contextual cues as in\ \cite{furnari2019rulstm}. This representation is essentially an object-centric feature that describes the attended environment (\ie where the camera wearer is facing towards), therefore 3D action location can not used for selecting surrounding environment features.

\noindent \textbullet\ \textbf{I3D+2DGround} projects the object information from the 3D environment to 2D ground plane. A similar representation is also considered in\ \cite{rhinehart2016learning}. Note that the predicted 3D action location will also be projected to 2D ground plane to select local environment features.

\noindent \textbullet\ \textbf{I3D+SemVoxel} is inspired by~\cite{zhou2018voxelnet}, where we use the semantic probabilistic distribution of all the vertices within each voxel as a feature descriptor. Therefore, the resulting environment representation is a 4D tensor with dimension $X \times Y \times Z \times C$, where $X$, $Y$, $Z$ represent the spatial dimension, and $C$ denotes the number of object labels from the 3D environment mesh annotation introduced in Sec.4. 

\noindent \textbullet\ \textbf{I3D+Affordance} follows~\cite{ego-topo} to use the afforded action distribution as feature descriptor for each voxel. The resulting representation is a 4D tensor with dimension $X \times Y \times Z \times N$, where $N$ denotes the number of action classes.The afforded action distribution is derived from the training set.

\noindent \textbf{Results}. Our results on the seen environment split is listed in Table~\ref{table:3DEnv}. Our method outperforms I3D-Res50 baseline by a large margin ($4.2\%/1.8\%$ on Mean Cls Acc/Top1 Acc) on action recognition. We attribute this significant performance gain to explicitly modeling the 3D environment context. As for 3D action localization, our method outperforms I3D-50 by $9.3\%$ -- a relative improvement of \textbf{69\%}. Notably, predicting the 3D action location based on video sequence alone is erroneous. Our method, on the other hand, explicitly models the 3D environment factor and thus improves the performance of 3D action localization. In subsequent sections, we will show that the performance improvement does not simply come from additional input modalities of 3D environment, but attributes to a careful design of 3D representation and probabilistic joint modeling.

\noindent \textbf{Comparison on environment representation}. We now compare HVR with other forms of environment representation. As shown in Table~\ref{table:3DEnv}, I3D+Obj has minor improvement on the over all performance, while I3D+2DGround, I3D+SemVoxel and I3D+Affordance can improve the performance of action recognition and 3D localization by a notable margin. Those results suggest that the environment context (even in 2D space) plays an important role in egocentric action understanding. More importantly, our method outperforms all previous methods by at least $1.7\%$ for action recognition and $5.3\%$ for 3D action localization. \emph{These results suggest that our proposed HVR is superior to a 2D ground plane representation}, and demonstrates that using the semantic occupancy map as the environment descriptor can better facilitate egocentric understanding.

\begin{table}[t]
\centering
\caption{Ablation study for the 3D representation. We present the results of our method that adopts different semantic occupancy map resolution $M$.}
{
\tablestyle{8pt}{1.0}
\begin{tabular}{c|cc|ccc}
\toprule
\multicolumn{1}{c|}{\multirow{3}{*}{Method}}          
&\multicolumn{2}{c|}{Action Recognition} &\multicolumn{3}{c}{3D Action Localization} \\ \cline{2-6}
& Mean Cls Acc &Top-1 Acc & Prec & Recall & F1  \\ \hline 
I3D-Res50    &37.48 &55.15 &8.14  &38.73   &13.45 \\
I3D+SemVoxel   &39.23 &56.07    &11.26  &38.77 &17.45 \\ 
Ours ($M=2$)      &39.04 &56.26  &12.19  &36.82 &18.32\\ 
Ours ($M=4$)        & {\fontseries{b}\selectfont 41.64} & {\fontseries{b}\selectfont 56.94} &{\fontseries{b}\selectfont 16.71} &35.55 &22.73\\
Ours ($M=8$)         &40.06 &56.04  &16.13 & {\fontseries{b}\selectfont 39.84} &{\fontseries{b}\selectfont 22.96} \\ 
\bottomrule
\end{tabular}}
\label{table:ablation}
\end{table}

\subsection{Ablation Studies} 

We now present detailed ablation studies of our method on seen split. To begin with, we analyze the role of semantic and geometric information in our hierarchical volumetric representation (HVR). We then present an experiment to verify whether fine-grained environment context is necessary for egocentric action understanding. Furthermore, we show the benefits of probabilistic joint modeling of action and 3D action location. 

\noindent \textbf{Semantic Meaning and 3D Geometry}.\ The semantic occupancy map carries both geometric and semantic information of the local environment. To show how each component contributes to the performance boost, we compare Ours with I3D+SemVoxel, where only semantic meaning is considered, in Table~\ref{table:ablation}. Ours outperforms I3D+SemVoxel by a notable margin for action recognition and a large margin for 3D localization. These results suggest that semantic occupancy map is more expressive than only semantic information for action understanding, yet it has smaller impact on action recognition than 3D action localization.

\noindent \textbf{Granularity of 3D Information}.\ We further show what level of 3D environment granularity is needed for egocentric action understanding. By the definition of occupancy map, increasing the resolution $M$ of children voxels will approximate the actual 3D shape of the environment. Therefore, we report results of our method with different occupancy map resolution in Table~\ref{table:ablation}. Not surprisingly, low occupancy map resolution lags behind Ours for action recognition by $2.6\%\downarrow$ and 3D action localization by $4.4\%\downarrow$, which again shows the necessity of incorporating the 3D geometric cues. Another interesting observation is that higher resolution can slightly increase the 3D action localization accuracy by $0.2\%$, yet decreases the performance on action recognition by $1.6\%\downarrow$. These results suggest that fine-grained 3D shape of the environment is not necessary for action recognition. In fact, higher resolution will dramatically increase the feature dimension of the environment representation, and thereby incurs more barriers to the network.

\begin{table}[t]
\footnotesize 
\centering
\caption{Ablation study for joint modeling of action category and 3D action location. Our proposed probabilistic joint modeling can consistently benefit the performance on action recognition and 3D action localization}
{
\setlength{\tabcolsep}{2.7pt} 
\tablestyle{8pt}{1.0}
\begin{tabular}{c|cc|ccc}
\toprule
\multicolumn{1}{c|}{\multirow{3}{*}{Method}}          
&\multicolumn{2}{c|}{Action Recognition} &\multicolumn{3}{c}{3D Action Localization} \\ \cline{2-6}
 & Mean Cls Acc &Top-1 Acc & Prec & Recall & F1  \\ \hline 
I3D-Res50    &37.48 &55.15 &8.14  &{\fontseries{b}\selectfont 38.73} &13.45 \\
I3D+GlobalEnv          &35.99  &54.93   &8.82  &36.40  &14.20\\	
I3D+DetEnv       &39.37  &55.88  &14.11   &32.66 &19.71\\ 
Ours       & {\fontseries{b}\selectfont 41.64}  &{\fontseries{b}\selectfont 56.94} &{\fontseries{b}\selectfont 16.71} &35.55 &{\fontseries{b}\selectfont 22.73} \\ 
\bottomrule
\end{tabular}}
\label{table:ablation:modeling}
\end{table}


\noindent \textbf{Joint Learning of Action and 3D Location}. We denote a baseline model that directly fuses global environment features, extracted by the same 3D convolutional network from our method, with video features for activity reasoning as I3D+GlobalEnv. The results are presented in Table\ \ref{table:ablation:modeling}. I3D+GlobalEnv decreases the performance of I3D-Res50 backbone network by $1.5\%\downarrow/0.2\%\downarrow$ for action recognition and has marginal improvement for 3D action localization ($+0.8\%$). We speculate that this is because only 3 types of scene reconstruction available for training may lead to overfitting. In contrast, our method makes use of the learned 3D action location to select interesting environment features associated with the action. As the action location varies among different input videos, our method can utilize the 3D environment context without running into the pitfall of overfitting, and therefore outperforms I3D+GlobalEnv by $5.7\%/2.0\%$ for action recognition and $8.5\%$ for 3D action localization.

\noindent \textbf{Probabilistic Modeling of 3D Action Location}. As introduced in Sec.4, considerable uncertainty lies in the prior distribution of 3D action location, due to the challenging artifact of 2D to 3D camera registration. To verify that the probabilistic modeling can account for the  uncertainty of 3D action location ground truth, we compare our method with a deterministic version of our model, denoted as DetEnv. DetEnv adopts the same inputs and network architecture as our method, except for the differentiable sampling with Gumbel-Softmax Trick. As shown in Table\ \ref{table:ablation:modeling}, Ours outperforms DetEnv by $2.3\%$ for action recognition and $3.0\%$ for 3D action localization. These results demonstrate the benefits of the stochastic units adopted in our method. 

\noindent \textbf{Remarks}. To summarize, our key finding is that both 3D geometric and semantic contexts convey important information for action recognition and 3D localization. Another important take home is that egocentric understanding only requires a sparse encoding of geometric information. Moreover, without a careful model design, the 3D environment representation has minor improvement on (or even decreases) the overall performance as reported in Table~\ref{table:ablation:modeling}.



\subsection{Generalization to Novel Environment}
\begin{table}[t]
\centering
\caption{Experimental results on unseen environment split. Our model show the capacity of better generalizing to an unseen environment with known 3D map. The best results are highlighted with \textbf{boldface}, and the second-best results are \underline{underlined}.}
{
\setlength{\tabcolsep}{2.7pt} 
\tablestyle{8pt}{1.0}
\begin{tabular}{c|cc|ccc}
\toprule
\multicolumn{1}{c|}{\multirow{3}{*}{Method}}          
&\multicolumn{2}{c|}{Action Recognition} &\multicolumn{3}{c}{3D Action Localization} \\ \cline{2-6}
& Mean Cls Acc &Top-1 Acc  & Prec & Recall & F1  \\ \hline 

I3D-Res50   &29.24 &52.22  &6.20  &45.14   &10.90 \\
I3D+2DObject       &29.91 &53.05    &6.31  &42.22  &10.98 \\ 
I3D+2DGround       &30.06 &\underline{53.87}    &6.95  &41.27  &11.90 \\ 
I3D+SemVoxel       &\underline{30.19} &53.37    &\underline{7.03}  &\underline{43.55}  &\underline{12.11} \\ 
Ours      & {\fontseries{b}\selectfont 31.55}  &{\fontseries{b}\selectfont 55.33}  &{\fontseries{b}\selectfont 7.50} & {\fontseries{b}\selectfont 44.97} &{\fontseries{b}\selectfont 12.86} \\ 
\bottomrule
\end{tabular}}
\label{table:unseen}
\end{table}

We further present experiment results on the unseen environment split in Table\ \ref{table:unseen}. Our model outperforms all baselines by a notable margin on both action recognition and 3D action localization. Note that the affordance map requires the observation of action instances on the 3D spatial location and thus cannot be applied on the unseen split. These results suggest that explicitly modeling the 3D environment context can improve the generalization ability to unseen environments with known 3D maps. However, the performance gap is smaller in comparison to the performance boost on seen split. We speculate that this is because we only have two different types of environments for training and therefore the risk of overfitting on unseen split is further exemplified.

\subsection{Discussion}
\begin{figure*}[t]
\centering
\includegraphics[width=1.0\linewidth]{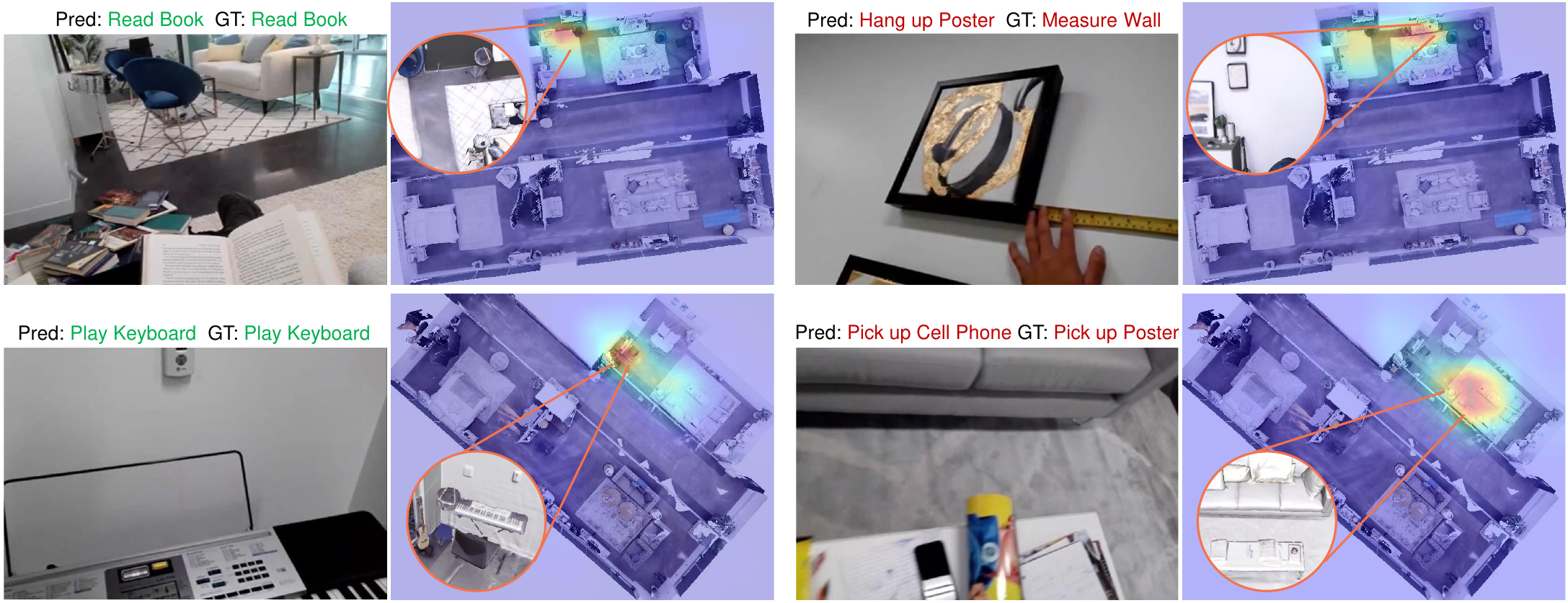}
\caption{Visualization of predicted 3D action location (projected on top-down view of the reconstructed 3D scene) and action labels (captions above the video frames). We present both {\color{green} successful} and {\color{red} failure} examples. We also show the ``zoom-in" spatial region of the action location to help readers to better interpret our action localization results.} 
\label{fig:vis}
\end{figure*}
\noindent\textbf{Visualization of Action Location}.\ We visualize our results on seen environment split. Specifically, we project the 3D saliency map of action location on the top-down view of the 3D environments. As shown in Fig.\ \ref{fig:vis}, our model can effectively localize the coarse action location and thereby select the region of interest from the global environment features for action recognition. By examining the failure cases, we found that the model may run into the failure modes when the video features are not sufficiently discriminative (\ie when the camera wearer is standing close to a white wall.)

\noindent\textbf{Limitation and Future Work}.\ One limitation of our method is the requirement of high-quality 3D reconstruction with object annotations. However, we conjecture that 3D object detection algorithms~\cite{zhou2018voxelnet}, semantic structure from motion~\cite{kundu2014joint} and 3D scene graphs~\cite{Armeni_2019_ICCV} can be used to replace the human annotation, since our current volumetric representation only adopts a low resolution semantic occupancy map as environment descriptor. We plan to explore this direction as our future work. Another limitation is the potential error in 2D to 3D camera registration, as discussed in Sec.4. Currently, only camera poses from key video frames can be robustly estimated. Our method thus does not model the location shift within the same action. We argue that camera registration can be drastically improved with the help of additional sensors (\eg IMU or depth camera). Incorporating those sensors into egocentric capturing setting is an exciting future direction. In addition, our method did not consider the camera orientation. We will leave this for future efforts.
 


\section{Conclusion}
We introduced a deep model that makes use of egocentric videos and a 3D map to address the novel task of joint action recognition and 3D localization. Our key insight is that the 3D geometric and semantic context of the surrounding environment provides critical information that complements video features for action understanding. The key innovation of our model is to characterize the 3D action location as a latent variable, which is used to select the surrounding local environment features for action recognition. Our model demonstrated impressive results on seen and unseen environments when evaluated on the newly released Ego4D dataset~\cite{grauman2021ego4d}. We believe our work provides a critical first step towards understanding actions in the context of a 3D environment, and points to exciting future directions in connecting egocentric vision and 3D scene understanding for AR and HRI. 

\smallskip
\noindent \textbf{Acknowledgments}. Portions of this project were supported in part by a gift from Facebook.
\clearpage
%
%
\bibliographystyle{splncs04}
\bibliography{egbib}

\newpage

This is the supplementary material for our ECCV 2022 paper, titled ``Egocentric Activity Recognition and Localization on a 3D Map''. The contents are organized as follows.
\begin{itemize}
\item \hyperref[sec:s1]{A} Implementation Details.
\item \hyperref[sec:s2]{B} Dataset Details.
\item \hyperref[sec:s3]{C} Analysis of Table 1 in Main Paper
\item \hyperref[sec:s4]{D} Additional Qualitative Results.
\item \hyperref[sec:s5]{E} Code and Licenses.
\end{itemize}

\renewcommand\thesection{\Alph {section}}
\renewcommand\thesubsection{\thesection.\arabic{subsection}}
\setcounter{section}{0}
\renewcommand*{\theHsection}{chX.\the\value{section}}

\section{Implementation Details}
\label{sec:s1}
\noindent\textbf{Data Processing}. We resize all video frames to the short edge size of 256. For the coarse 3D map, we adopt a resolution of $28\times 28 \times 8$ for parent voxel, and $M=4$ for children voxels. For training, our model takes an input of 8 frames (temporal sampling rate of 8) with a resolution of $224\times224$. For inference, our model samples 30 clips from a video (3 along spatial dimension and 10 in time). Each clip has 8 frames with a resolution of $224\times224$. We average the scores of all sampled clips for video level prediction.\smallskip 

\noindent\textbf{Training Details}. Our model is trained using SGD with momentum 0.9 and batch size 64 on 4 GPUs. The initial learning rate is 0.0375 with cosine decay. We set weight decay to 1e-4 and enable batch norm~\cite{ioffe2009batch}. To avoid overfitting, we adopt several data augmentation techniques, including random flipping, rotation, cropping and color jittering. Our model is implemented in PyTorch. Our scource code is included with this supplement and will be made publicly available.

\section{Dataset Details}
\label{sec:s2}
This section introduces details of our new egocentric video dataset. We present additional qualitative results of the key frame camera registration, and show additional rendered images of the 3D reconstructions from our dataset. Finally, we plot the distribution of activity categories in our dataset.\smallskip

\begin{figure*}[t]
\centering
\includegraphics[width=0.9\linewidth]{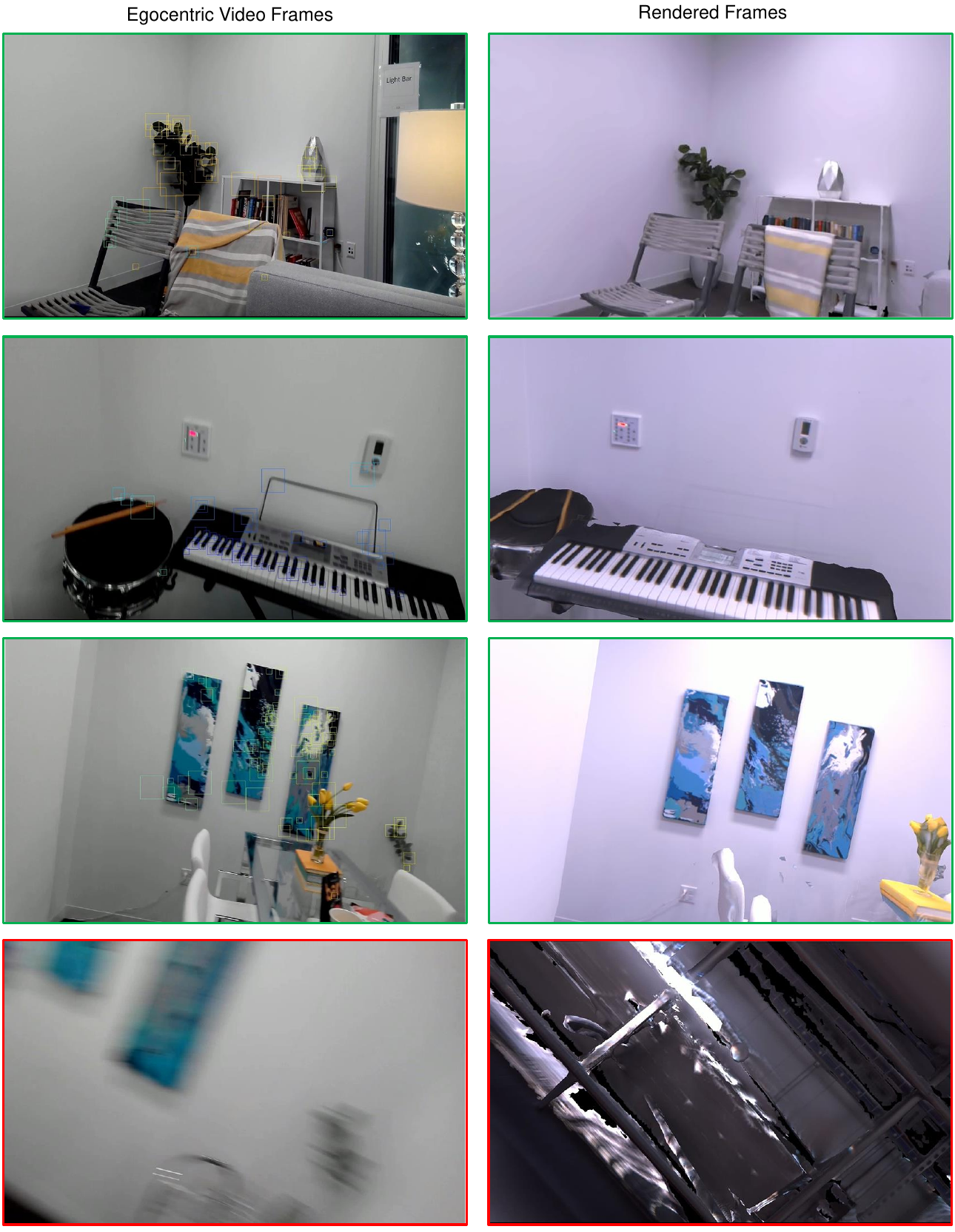}
\caption{Qualitative Results for Camera Registration. We present both {\color{green} successful} cases and {\color{red} failure} cases of the camera registration results by using the estimated camera pose to render the egocentric view of the 3D environment reconstructions.}
\vspace{-1em}
\label{fig:registration}
\end{figure*}

\noindent\textbf{Camera Registration}.\ We provide qualitative results of key frame egocentric camera registration in Fig.\ \ref{fig:registration}. Specifically, we use the estimated camera pose to render the egocentric view of the 3D environment. When the camera registration fails (\eg insufficient inliers are matched by RANSAC), we assign a dummy result at the world origin, which will result in a completely wrong rendered view of the 3D environment as in the last row of Fig.~\ref{fig:registration}.  We also visualize the matched feature points~\cite{alahi2012freak} to help readers better interpret the RANSAC based matching method for camera registration introduced in Sec.\ 4 of the main paper. Notably, registering the egocentric camera into a 3D environment based on only images remains a major challenge. The motion blur, foreground occlusion, change of illumination, and featureless surfaces in indoor capture are the main failure causing factors of the 2D-to-3D registration. To mitigate these failure cases, we only consider camera relocalization using key frames to approximate the ground truth of 3D action locations, and do not model how location might shift over time within the same action. Note that, if the entire action clips does not have one robust key frame registration result, we adopt uniform distribution for the 3D action location prior $q(r|x,e)$ during training.\smallskip

\begin{figure*}[t]
\centering
\includegraphics[width=0.95\linewidth]{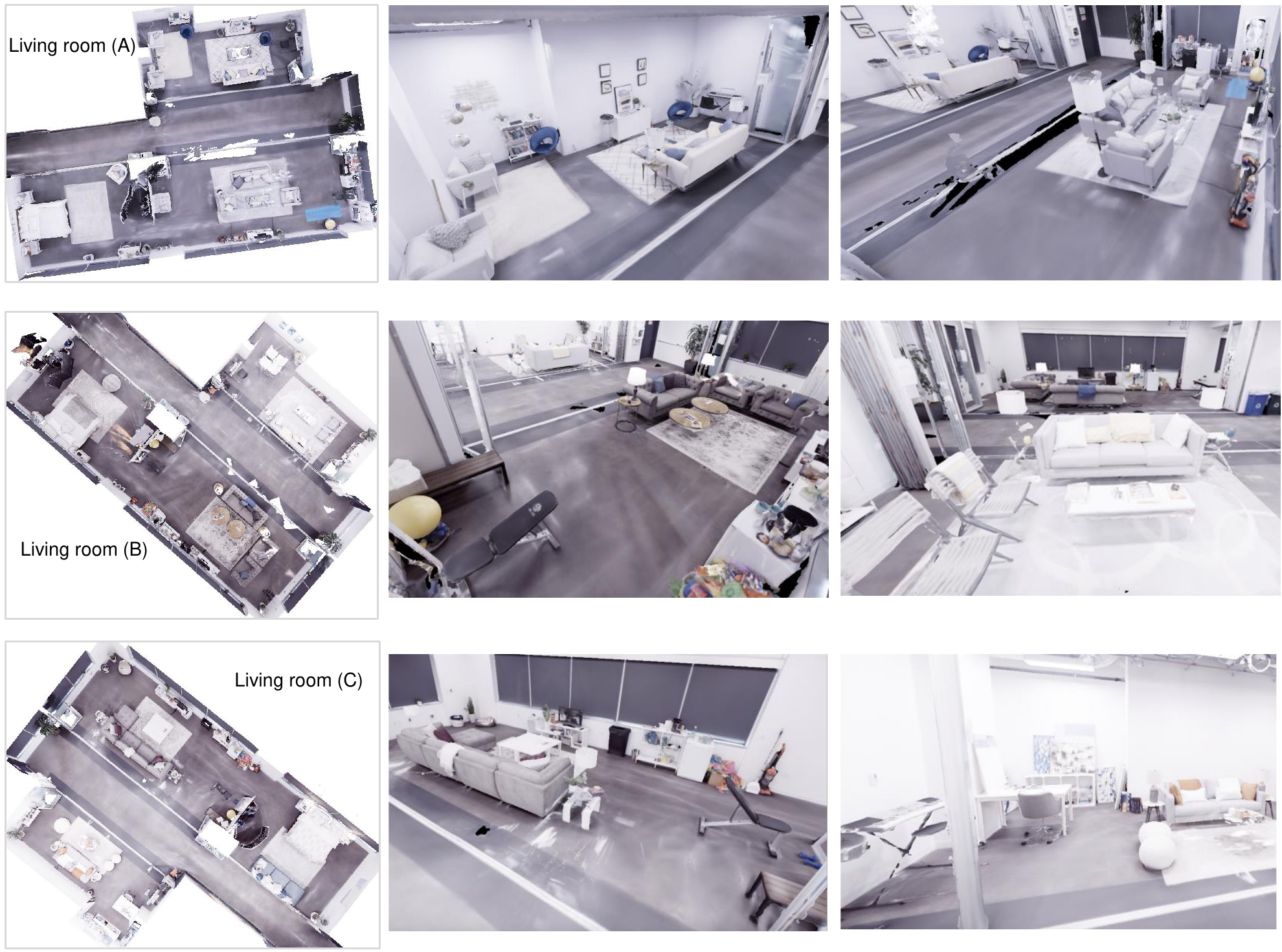}
\caption{Rendered views of the 3D environment reconstructions captured in our dataset.}

\label{fig:3Dscene}
\end{figure*}

\noindent\textbf{3D Environment Reconstruction}.\ Our dataset captures the 3D reconstructions of three different living rooms using SOTA dense Reconstruction system~\cite{replica19arxiv}. We provide more rendered images of the 3D reconstructions in Fig.\ \ref{fig:3Dscene}.\smallskip

\begin{figure*}[t]
\centering
\includegraphics[width=0.95\linewidth]{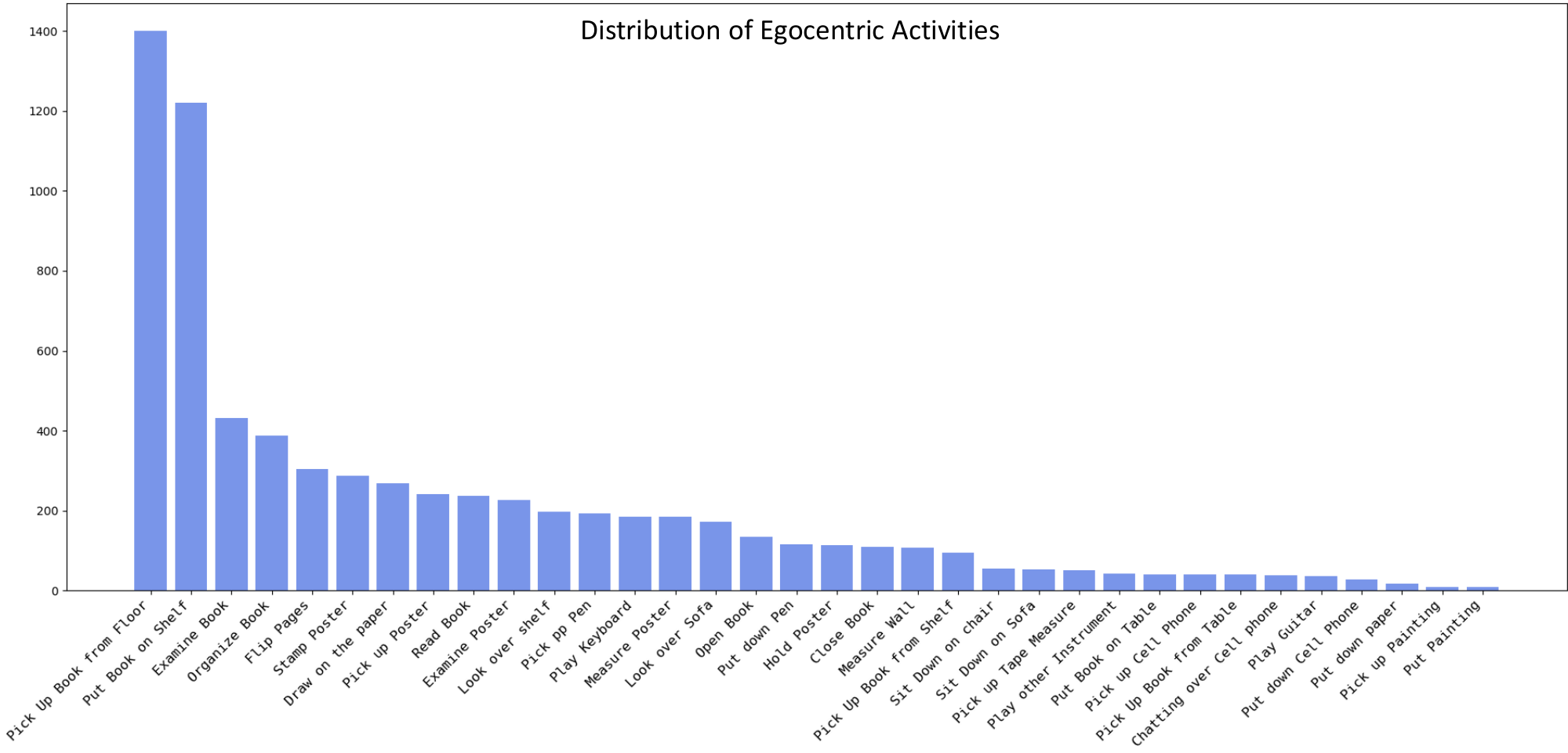}
\caption{Long tailed activity distribution in our dataset.}
\label{fig:distribution}
\end{figure*}

\noindent\textbf{Activity Distribution}.\ We present the distribution of egocentric activities in our dataset in Fig.\ \ref{fig:distribution}. Similar to~\cite{Damen2018EPICKITCHENS,li2020eye}, our dataset has a ``long tailed'' distribution that characterizes naturalistic human behavior. For example, the action activity category of ``Pick up Book from Floor'' happens 1400 times, while the action of ``Put Painting'' on the tail occurs only 9 times. Mean Class Accuracy provides a metric that is not biased towards frequently occurred categories, and thus is better suited than Top-1 accuracy for activity recognition on our dataset. We highlight that our model outperforms I3D baseline by $4.2\%$ on Mean Class Accuracy.

\begin{figure*}[t]
\centering
\includegraphics[width=0.9\linewidth]{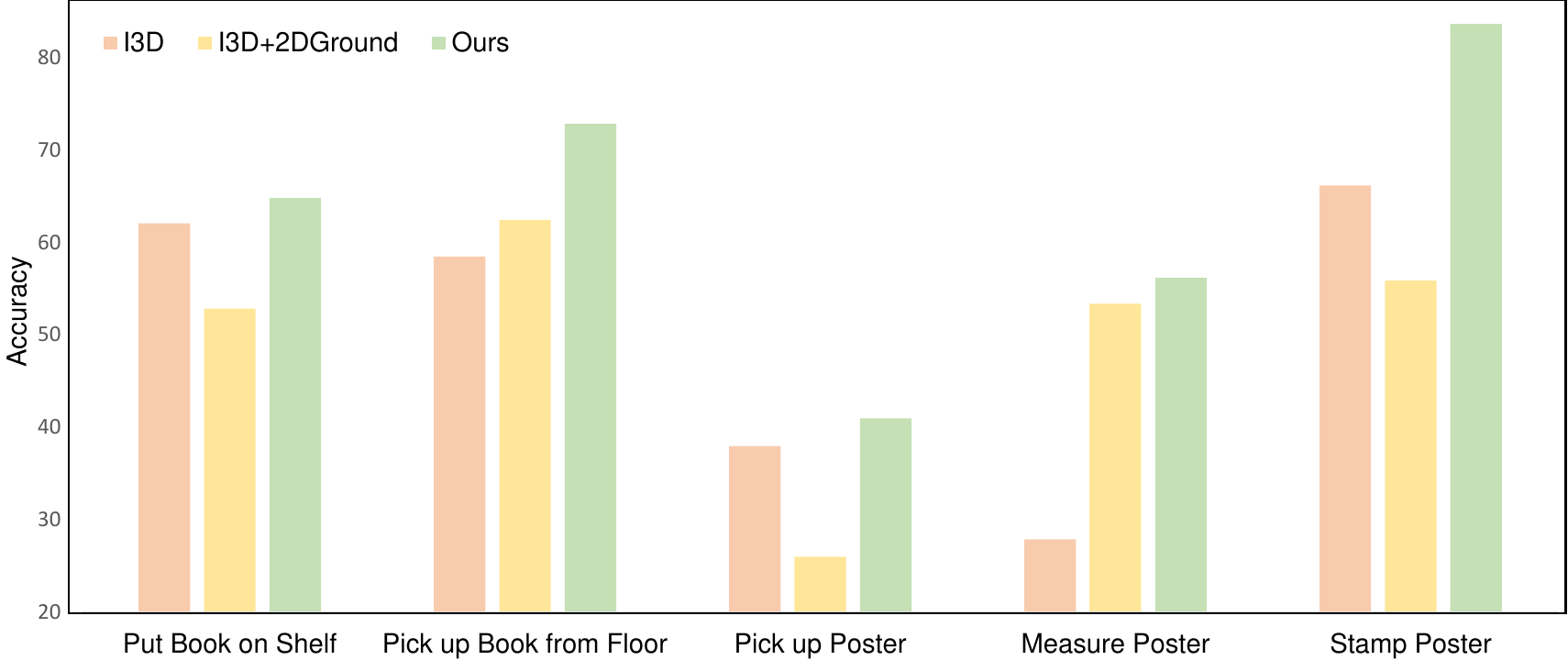}
\caption{A closer look at the experiment results comparison in Table 1 of main paper.}
\label{fig:bargragh}
\end{figure*}

\section{Analysis of Table 1 in Main Paper}
\label{sec:s3}

\noindent\textbf{Remarks on Environment Representation in Table 1}.\ Although our method requires the annotation of static objects, we are the first to show how the 3D scene proximity can facilitate egocentric video understanding. Moreover, baseline methods like I3D+SemVoxel and I3D+2DGround both use the same static objects annotation to describe the environment context as our method. Therefore, it is a fair experiment comparison between our approach and those methods. I3D+Affordance also requires extra inputs, in the form of afforded action distribution across the entire 3D map, which directly links each voxel in the map to the most likely action for that location, a much stronger prior than our HVR representation. Note that the affordance map requires that the observation of action instances densely cover the full 3D scene (which is why I3D+Affordance cannot be applied on the unseen split). Therefore, we argue that I3D+Affordance is less scalable than our model.\smallskip

\noindent\textbf{Baseline Details in Table 1}.\ I3D+SemVoxel/+Affordance adopt the same architecture as our model, and the only difference is the number of input features. I3D+2DObj adopts the late fusion strategy as~\cite{furnari2019rulstm}. I3D+2DGround adopts 2D convolutional operations for extracting 2D environment features. The extracted features are further tiled into a 4D tensor and concatenated with video features for recognition. We will add those descriptions in camera ready.\smallskip

\noindent\textbf{Additional Analysis}.\ We specifically compare the activity classes, where our model significantly outperforms I3D and I3D+2DGround in Fig.~\ref{fig:bargragh}. Interestingly, our model performs better at classes where the video features might not be discriminative enough for recognition, \eg ``Pick up Book from Floor'' vs.\ ``Put Book on Shelf'', or ``Pick up Poster'' vs.\ ``Stamp Poster''. Moreover, the contextual features from 2D ground plane provide limited information for understanding those egocentric activities. We conjecture that our method makes use of environment features surrounding the predicted 3D action location to complement video features for activity recognition.

\begin{figure*}[t]
\centering
\includegraphics[width=0.95\linewidth]{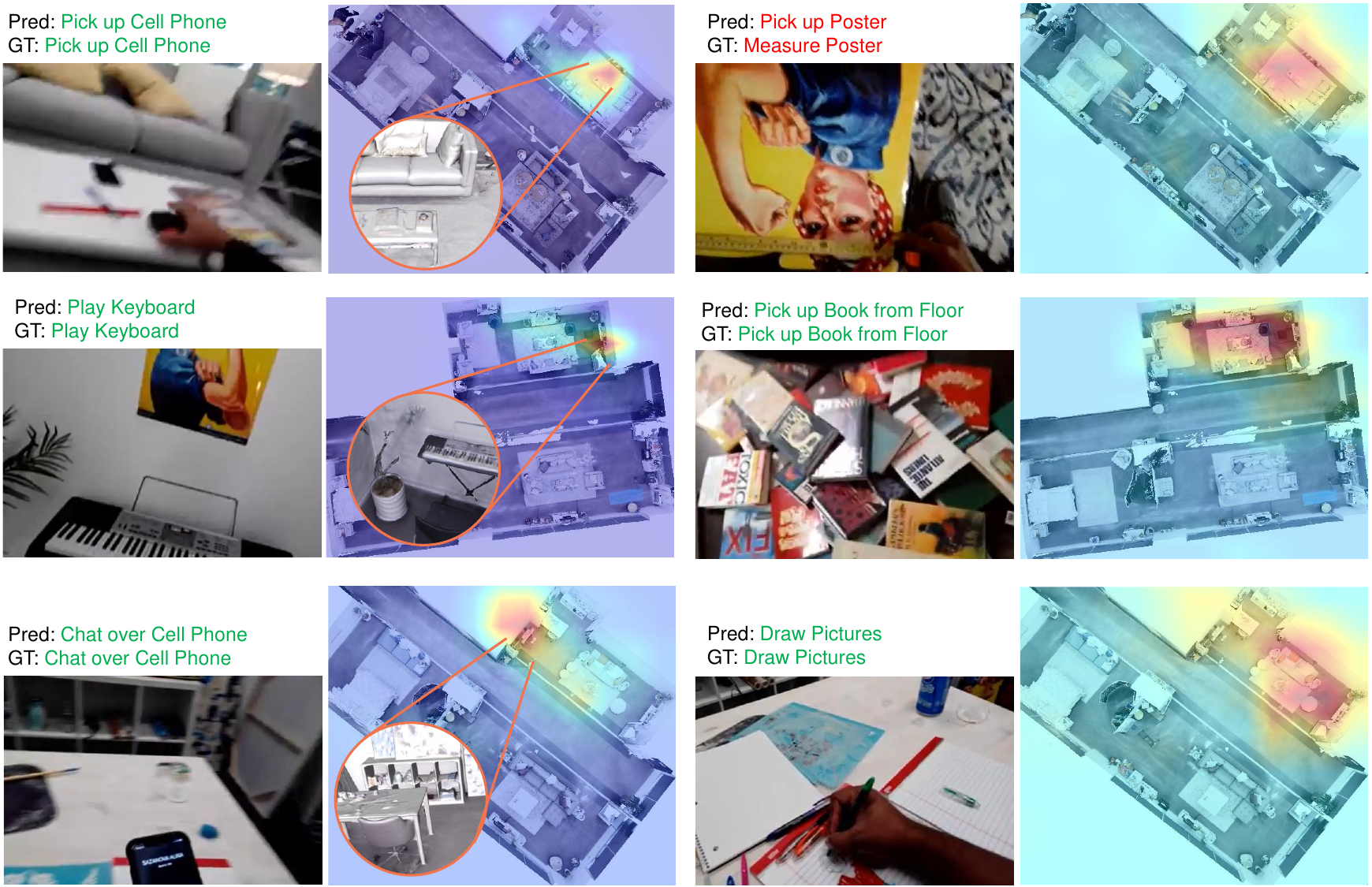}
\caption{Additional visualization of predicted 3D action location and action labels.}
\label{fig:more_vis}
\end{figure*}

\section{Additional Qualitative Results}
\label{sec:s4}
Finally, we provide additional qualitative results. As shown in Fig.\ \ref{fig:more_vis}, we present predicted 3D action location and action labels. The figure follows the same format as Fig.~3 in the main paper. Those results suggest that our model can effectively localize the action location and thereby more accurately predict the action labels. Another interesting observation is that the model may output a ``diffused'' heatmap, when the foreground active objects take up the majority of the video frames (right column of Fig.\ \ref{fig:more_vis}). This is because the model receives uniform prior as supervisory signals when the camera registration fails for an action clip. In these cases, our model opts for predicting a diffused heat map of action location to prevent itself from missing important environment features. In doing so, our model might still be able to successfully predict the action labels, despite the failure of camera registration.

\section{Code and Licenses}
\label{sec:s5}
Our implementation is built on top of~\cite{fan2020pyslowfast}, which is under the Apache License\footnote{\url{https://github.com/facebookresearch/SlowFast/blob/main/LICENSE}}. 

\begin{table*}[t]
\caption{Network architecture of our two-pathway network. We omit the residual connection in backbone ResNet-50 for simplification. }
\label{table:structure}
\small
\def\arraystretch{1.36}
\setlength{\tabcolsep}{0.3pt}
\centering
\scalebox{0.75}{

\begin{tabular}{c|c|c|c|c|c|c}
\hline 
\multirow{2}{*}{\textbf{ID}} & \multirow{2}{*}{\textbf{Branch}}                                                       & \multirow{2}{*}{\textbf{Type}}                                                & \multirow{2}{*}{\begin{tabular}[c]{@{}c@{}}\textbf{Kernel Size}\\ THW,(C)\end{tabular}} & \multirow{2}{*}{\begin{tabular}[c]{@{}c@{}}\textbf{Stride}\\ THW\end{tabular}} & \multirow{2}{*}{\begin{tabular}[c]{@{}c@{}}\textbf{Output Size}\\ THWC\end{tabular}} & \multirow{2}{*}{\textbf{Comments (Loss)}}                                                  \\
&   &   &   &   &   &                                                                                    \\ \hline 
1   & \multirow{24}{*}{\begin{tabular}[c]{@{}c@{}}Backbone\\ Input Size:\\ 8x224x224x3\end{tabular}} 
& Conv3D    & 5x7x7,64     & 1x2x2     & 8x112x112x64 \\ \cline{3-7}
2  & & MaxPool1  & 1x3x3  & 1x2x2  & 8x56x56x64  &   \\ \cline{3-7}
3   &    & \begin{tabular}[c]{@{}c@{}}Layer1\\Bottleneck 0-2 \end{tabular}    & \begin{tabular}[c]{@{}c@{}}3x1x1,64\\ 1x3x3,64\\1x1x1,256\end{tabular} ($\times$3)     & \begin{tabular}[c]{@{}c@{}}1x1x1\\ 1x1x1\\1x1x1\end{tabular} ($\times$3)     & 8x56x56x256  &  \\ \cline{3-7}
4    & & MaxPool2  & 2x1x1  & 2x1x1  & 4x56x56x256  &  \\ \cline{3-7}
5   &    & \begin{tabular}[c]{@{}c@{}}Layer2\\Bottleneck 0 \end{tabular}    & \begin{tabular}[c]{@{}c@{}}3x1x1,128\\ 1x3x3,128\\1x1x1,512\end{tabular}     & \begin{tabular}[c]{@{}c@{}}1x1x1\\ 1x2x2\\1x1x1\end{tabular}     &   &  \\ \cline{3-7}
6   &    & \begin{tabular}[c]{@{}c@{}}Layer2\\Bottleneck 1-3 \end{tabular}    & \begin{tabular}[c]{@{}c@{}}3x1x1,128\\ 1x3x3,128\\1x1x1,512\end{tabular}  ($\times$3)    & \begin{tabular}[c]{@{}c@{}}1x1x1\\ 1x2x2\\1x1x1\end{tabular} ($\times$3) & 4x28x28x512 & \begin{tabular}[c]{@{}c@{}} Concat with Env Feat for\\  3D Action Location Prediction\end{tabular}   \\ \cline{3-7}
7   &    & \begin{tabular}[c]{@{}c@{}}Layer3\\Bottleneck 0 \end{tabular}   & \begin{tabular}[c]{@{}c@{}}3x1x1,256\\ 1x3x3,256\\1x1x1,1024\end{tabular}     & \begin{tabular}[c]{@{}c@{}}1x1x1\\ 1x2x2\\1x1x1\end{tabular}   &  &  \\ \cline{3-7}
8   &    & \begin{tabular}[c]{@{}c@{}}Layer3\\Bottleneck 1-5 \end{tabular}   & \begin{tabular}[c]{@{}c@{}}3x1x1,256\\ 1x3x3,256\\1x1x1,1024\end{tabular}  ($\times$5)     & \begin{tabular}[c]{@{}c@{}}1x1x1\\ 1x1x1\\1x1x1\end{tabular}   ($\times$5)  &4x14x14x1024  &  \\ \cline{3-7}

9   &    & \begin{tabular}[c]{@{}c@{}}Layer4\\Bottleneck 0 \end{tabular}    & \begin{tabular}[c]{@{}c@{}}3x1x1,128\\ 1x3x3,128\\1x1x1,512\end{tabular}     & \begin{tabular}[c]{@{}c@{}}1x1x1\\ 1x2x2\\1x1x1\end{tabular}     &  &  \\ \cline{3-7}
10   &    & \begin{tabular}[c]{@{}c@{}}Layer4\\Bottleneck 1-2 \end{tabular}    & \begin{tabular}[c]{@{}c@{}}3x1x1,128\\ 1x3x3,128\\1x1x1,512\end{tabular}  ($\times$2)    & \begin{tabular}[c]{@{}c@{}}1x1x1\\ 1x2x2\\1x1x1\end{tabular} ($\times$2)     &4x7x7x2048  & \begin{tabular}[c]{@{}c@{}} Concat with Env Feat \\ for Activity Recognition\end{tabular}     \\ \cline{3-7} \hline

11 & \multirow{8}{*}{\begin{tabular}[c]{@{}c@{}}EnvNet \\ Input Size\\ 8x28x28x64\end{tabular}}       & \begin{tabular}[c]{@{}c@{}}Conv3d 1\end{tabular} & 3x3x3,356    & 2x1x1    & 4x28x28x256 &    \\ \cline{3-7}
12 &      & Conv3d 2   & 1x3x3,512 & 1x1x1  & 4x28x28x512 & \begin{tabular}[c]{@{}c@{}} Concat with Video Feat for\\  3D Action Location Prediction\end{tabular}   \\ \cline{3-7}
13 &      & \begin{tabular}[c]{@{}c@{}}Action Location Branch\\Conv3d 1\end{tabular}    & 1x3x3,512 &1x2x2 &  4x14x14x512 &  \\ \cline{3-7}
13 &      & \begin{tabular}[c]{@{}c@{}}Action Location Branch\\Conv3d 2\end{tabular}  & 1x3x3,1 &1x1x1 &  4x14x14x1 & KLD Loss \\ \cline{3-7}
14          &                                                                               & \begin{tabular}[c]{@{}c@{}}Gumbel Softmax 1\\ (Sampling)\end{tabular}                                                        &                                                                              &                                                                         & 4x14x14x1                                                                              &      Sampling 3D Action Location                                                                            \\ \cline{3-7} 
15 &      & Maxpool  & 2x1x1 &2x1x1 &  4x14x14x512 &  \\ \cline{3-7} 
16 &      & Conv3d 3   & 1x3x3,1024 & 1x1x1  & 4x14x14x1024  & \\ \cline{3-7}
17 &      & Weighted Avg Pooling  & & & 4x14x14x1024  & \begin{tabular}[c]{@{}c@{}}Guided by\\Sampled 3D Action Location\end{tabular} \\ \cline{3-7}
18 &      & Conv3d 4  &1x3x3,1024 &1x3x3 & 4x7x7x1024  & \begin{tabular}[c]{@{}c@{}} Concat with Video Feat \\ for Activity Recognition\end{tabular}    \\ \hline

22 & \multirow{4}{*}{\begin{tabular}[c]{@{}c@{}}Recognition \\ Network\end{tabular}}       & Weighted Avg Pooling & 4x7x7    & 4x7x7    & 1x1x1x1024 &   \begin{tabular}[c]{@{}c@{}} Fused Environmental \\ and Video features\end{tabular}  \\ \cline{3-7}
23                                      &                                                                               & Fully Connected                                                      &                                                                              &                                                                         & 1x1x1xN                                                                        &                                                                                  \\ \cline{3-7} 
24                                      &                                                                               & Softmax                                                              &                                                                              &                                                                         & 1x1x1xN                                                                & Cross Entropy Loss\\ \hline

\end{tabular}}
\end{table*}

\end{document}